# Automated Dental Caries Segmentation in Panoramic Radiographs Using Dual-Stage Deep Learning

Jihun Kim[1,2,*], Kyeonghun Kim[3,*], Jong-yeol Lee[4], Yeongseok Seo[5,†], and Dohyun Chun[6,†]

[1] College of Humanities & Social Sciences Convergence, Yonsei University, Wonju-si, Gangwon-do, Korea.
[2] Yonsei Institute of AI Data Convergence Science, Yonsei University, Wonju-si, Gangwon-do, Korea.
[3] OUTTA, Seoul, Korea.
[4] The One Star Co., Ltd., Uijeongbu-si, Gyeonggi-do, Korea.
[5] WaveString, Seoul, Korea.
[6] College of Business Administration, Kangwon National University, Chuncheon-si, Korea

**Abstract.** Early detection of dental caries remains challenging due to limitations in traditional diagnostic methods, particularly for proximal lesions in posterior teeth. Deep learning models show promise for automated caries detection but face scalability constraints due to requirements for large volumes of expertly annotated training data. This study presents a dual-stage deep learning framework combining Faster R-CNN for tooth localization with U-Net for pixel-wise caries segmentation in panoramic radiographs. We developed a systematic transformation pipeline to convert large-scale polygon-annotated datasets into high-resolution binary segmentation masks, enabling pixel-wise supervised learning. The framework was trained using both expert-verified datasets and algorithmically-processed labels from 3,000 panoramic images. Our approach achieved robust performance with IoU of 0.9013, Dice coefficient of 0.9482, Recall of 0.9433, and Precision of 0.9774, demonstrating superior accuracy compared to existing methods while significantly reducing false positive rates. The dual-stage framework effectively addresses data annotation bottlenecks in dental AI applications and demonstrates potential for scalable, automated caries detection systems that can improve diagnostic consistency and support clinical decision-making.

## 1 Introduction

Dental caries remains one of the most prevalent chronic diseases worldwide, affecting individuals across all age groups and contributing to significant oral health burdens [1, 2]. Early detection is critical for preventing irreversible destruction of dental hard tissue and enabling timely preventive interventions [3, 4]. However, early-stage lesions, particularly those in the proximal regions of posterior teeth, are often difficult to detect due to anatomical limitations and insufficient visibility [5-7]. Traditional caries detection relies heavily on visual-tactile examination and radiographic interpretation, both of which suffer from inherent limitations that compromise diagnostic accuracy [8, 9]. Visual-tactile methods demonstrate low sensitivity for incipient lesions and are subject to considerable subjectivity based on clinician experience [7, 8]. Radiographic interpretation varies considerably across clinicians, with studies reporting significant inter-rater variability in diagnostic outcomes [10, 11]. Although adjunctive technologies such as laser fluorescence-based detection devices have been introduced, their clinical adoption remains limited due to high costs and variable diagnostic performance [14-16].

---

† Yeongseok Seo and Dohyun Chun are corresponding authors and contributed equally to this work.
* Jihun Kim and Kyeonghun Kim contributed equally to this work as the first authors.

Recent developments in artificial intelligence and deep learning have enabled new diagnostic capabilities in dental radiology [17-20]. Convolutional neural networks (CNNs) have demonstrated remarkable effectiveness in various dental imaging applications, including caries detection from panoramic and bitewing radiographs [21-24]. U-Net-based models have shown particular promise for medical image segmentation tasks due to their ability to capture both local and global contextual information [25-28]. However, a critical challenge in implementing deep learning frameworks lies in the requirement for large volumes of expertly annotated training data, which creates a significant bottleneck in clinical environments [29]. Manual pixel-wise annotation of dental caries demands substantial domain expertise and time investment, severely limiting the scalability of supervised learning approaches.

This study presents a dual-stage deep learning framework that combines Faster R-CNN for tooth localization with U-Net for pixel-wise caries segmentation in panoramic radiographs [25, 32]. Our key contribution lies in developing a systematic transformation pipeline that converts large-scale polygon-annotated datasets into high-resolution binary segmentation masks, enabling pixel-wise supervised learning without the prohibitive costs of manual annotation. The framework integrates both expert-verified and algorithmically-processed labels, effectively bridging the gap between annotation availability and model performance requirements while substantially expanding the utility of existing datasets for advanced segmentation model training [33-35].

Experimental evaluation demonstrates that our approach achieves robust performance with IoU of 0.9013, Dice coefficient of 0.9482, and significantly reduced false positive rates compared to existing methods, highlighting the potential for scalable, automated caries detection systems that can improve diagnostic consistency and support clinical decision-making.

## 2 Material and methods

### 2.1 Dataset

This study employed two complementary datasets to train and evaluate the proposed dual-stage caries segmentation framework. The primary dataset comprised 597 panoramic radiographs collected from Zhejiang Provincial People's Hospital, annotated by five experienced dental experts. The dataset was partitioned into 497 images for training and 100 images for testing, with expert dentists manually delineating bounding boxes for individual teeth and pixel-wise binary masks for carious lesions.

The secondary dataset consisted of 3,000 panoramic radiographs obtained from AI Hub, a comprehensive AI platform operated by the National Information Society Agency under Korea's Ministry of Science and ICT. This collection was constructed through a government-led medical AI data construction initiative involving 11 participating hospitals. The dataset featured polygon-based annotations stored in JSON format, which required systematic preprocessing to convert into pixel-level segmentation masks.

The polygon-based annotations from the secondary dataset required systematic transformation into pixel-wise segmentation masks for U-Net training. The preprocessing pipeline converted vector-based annotations into dense supervision signals for semantic segmentation. The transformation process commenced with parsing metadata from JSON annotation files. Coordinate pairs defining lesion boundaries and bounding box coordinates for individual teeth were extracted. For each image, an empty binary mask array was initialized with dimensions identical to the original radiograph. Polygon coordinates were reformatted and rendered onto the binary mask, assigning pixel values of 255 to lesion interiors while preserving background pixels as zero.

Individual tooth-level regions were extracted using provided bounding box coordinates. Both RGB images and corresponding binary masks were cropped to generate focused sub-images of individual teeth. Conditional validation excluded crops with zero dimensions. Quality control was implemented through visual cross-validation by superimposing processed masks onto source images, confirming preservation of location and shape without spatial distortions.

The processed secondary dataset served as secondary supervised training following initial training on the Hospital A expert-annotated dataset. This dual-stage approach utilized 597 expert-verified images for primary training, then 3,000 algorithmically-processed images for secondary training, transforming polygon annotations into 18,131 individual tooth-level training samples.

Images were cropped to focus on the oral cavity (20–80% width, 20–90% height), and CLAHE was applied to improve contrast. Class imbalance was addressed by randomly selecting equal numbers of healthy tooth samples to match the 635 carious samples. Augmentation techniques including flipping, rotation, and brightness adjustment were applied, resulting in training sets ranging from 1,270 to 7,620 images (Table 1). This study was conducted with exemption from ethical review as confirmed by the relevant Institutional Review Board (IRB exemption number: P01-202503-01-041).

**Table 1.** Summary of the number of images used for each model in the training and test sets.

| Model | Training set | Test set |
|---|---|---|
| Faster R-CNN | 90 | 10 |
| U-Net | 2729 | 295 |
| U-Net (balanced) | 1270 | 295 |
| U-Net (balanced with augmented) | 7620 | 295 |

## 2.2 Network Architecture

The proposed framework combines Faster R-CNN for tooth detection and U-Net for caries segmentation, as illustrated in Fig. 1. Faster R-CNN localizes individual teeth within panoramic radiographs, while U-Net performs pixel-level segmentation of carious lesions within detected tooth regions.

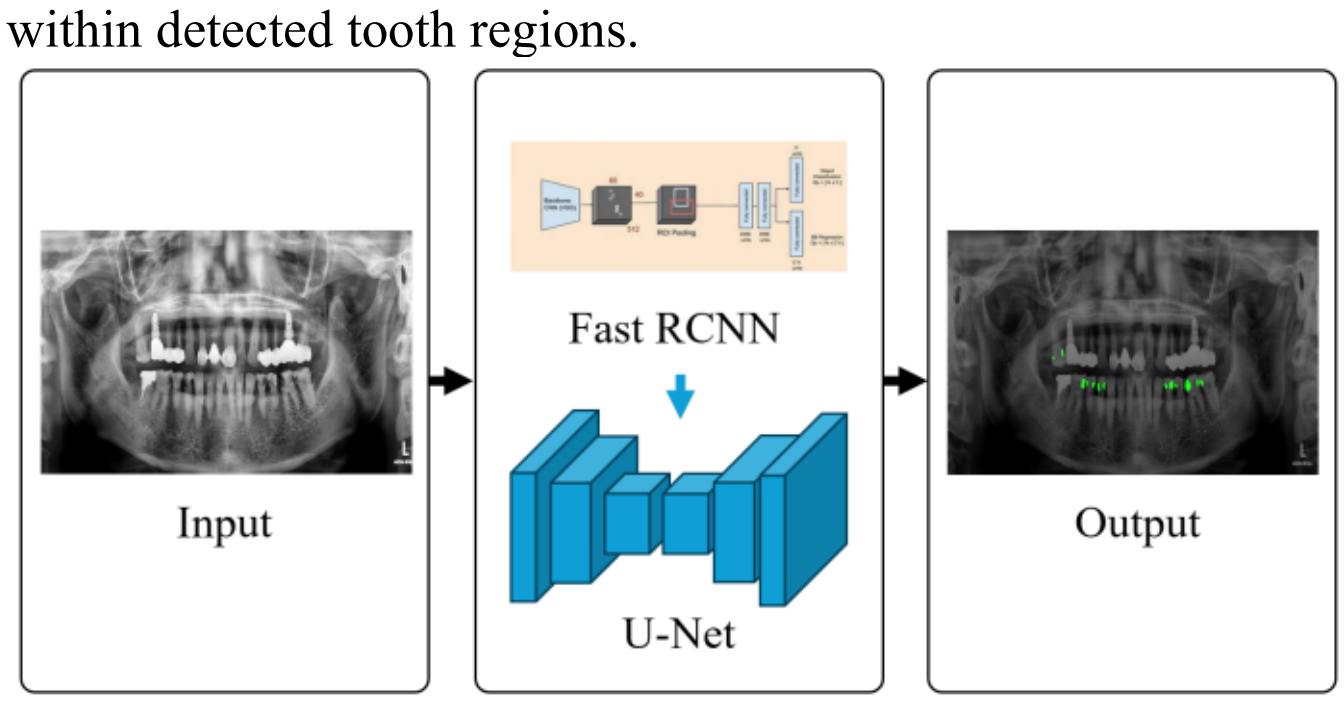

**Fig. 1** The overall process of the proposed method for dental caries segmentation.

Faster R-CNN was configured with MobilenetV3 as the feature extractor backbone (Fig. 2). MobilenetV3 employs depthwise separable convolutions and inverted residuals to reduce computational cost while preserving accuracy. The model was initialized with ImageNet pre-trained weights. The Region Proposal Network generated anchors of various sizes and aspect ratios to detect teeth of different shapes.

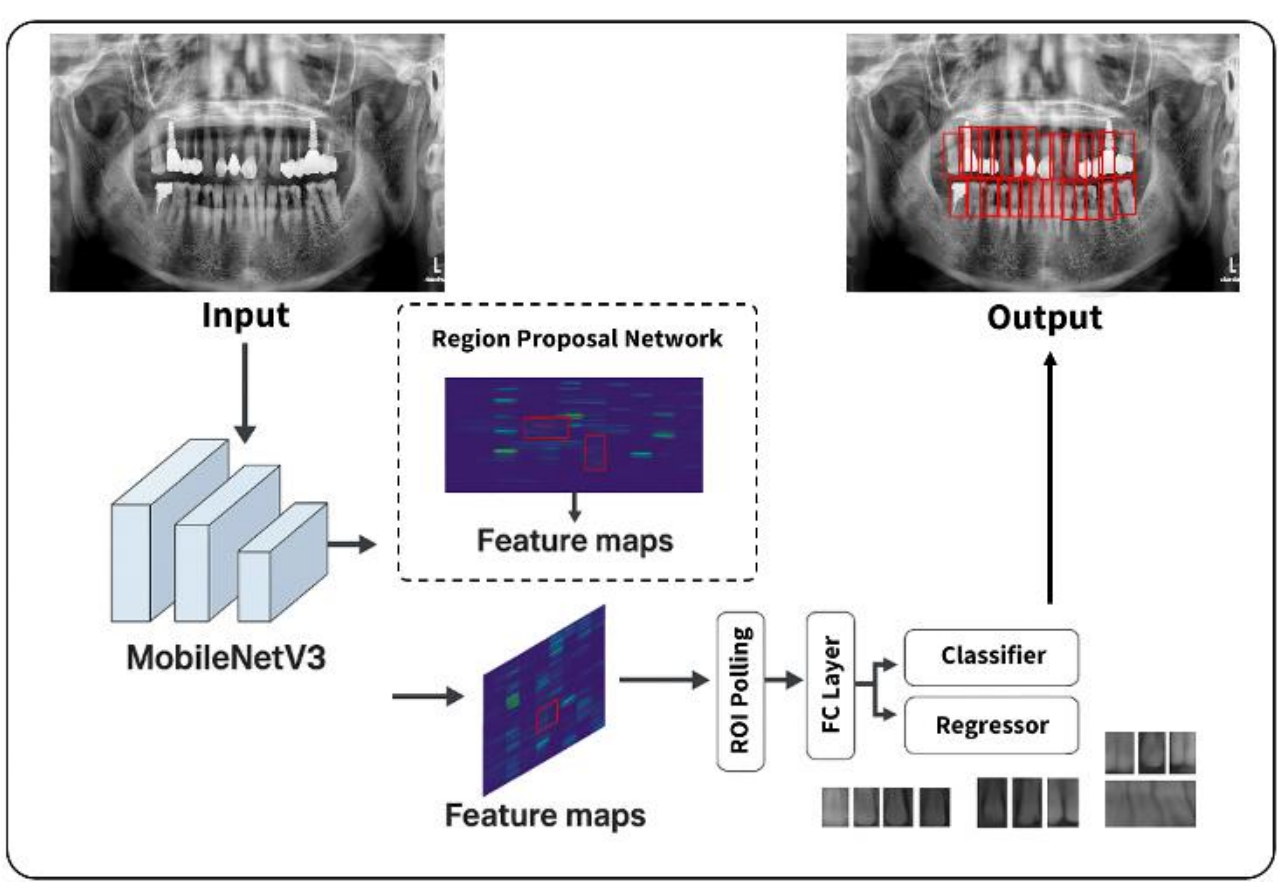

**Fig. 2** The architecture of Faster R-CNN with MobilenetV3 as the backbone, illustrating an example of tooth detection and the result of cropping the image using bounding box coordinates.

U-Net was adopted for lesion segmentation with an EfficientNet-B0 encoder (Fig. 3). EfficientNet-B0 uses compound scaling and Mobile Inverted Bottleneck Convolution blocks. Tooth images were resized to 256×256 pixels before being input into the model. The output was a binary mask indicating carious regions.

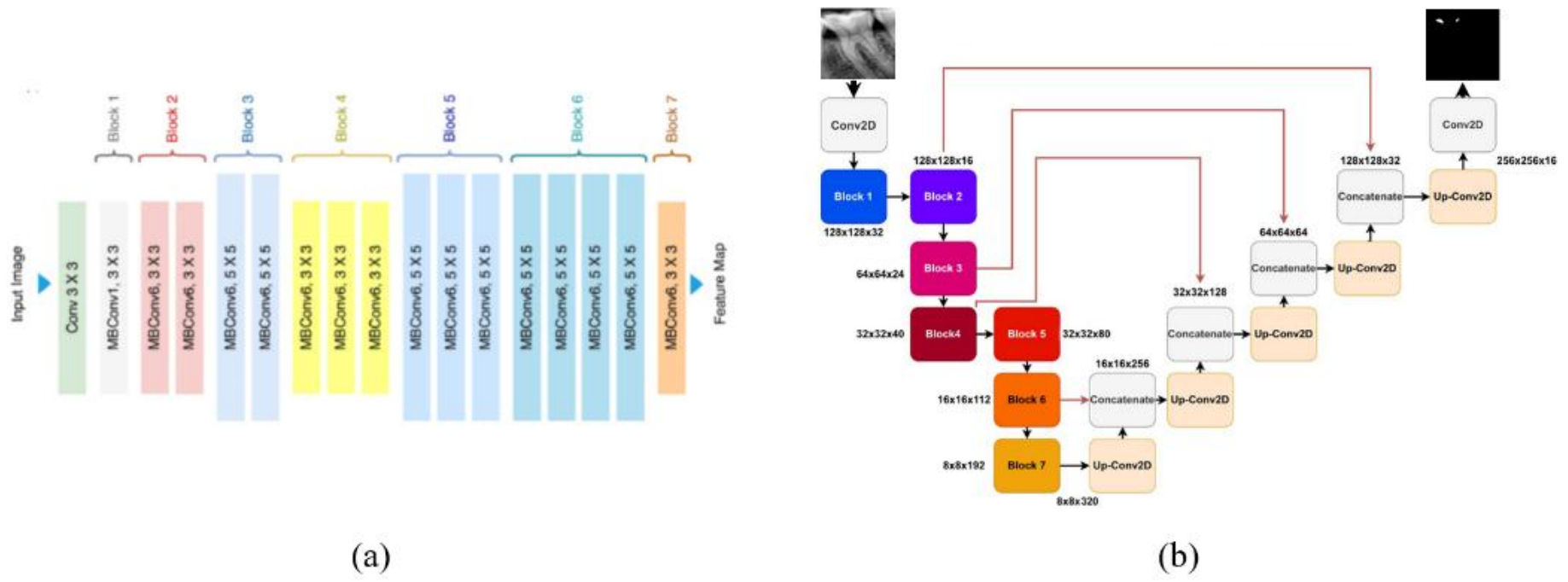

**Fig. 3**. Overview of the Network Architectures. (a) Block representation of EfficientNet-B0 Architecture, (b) Schematic of the U-Net architecture with an EfficientNet-B0 encoder.

## 2.3 Training Configuration and Evaluation

Post-processing procedures integrated individual U-Net predictions into panoramic-level segmentation outputs. Binary masks generated at 256×256 resolution were resized proportionally to match bounding box coordinates. A composite output was assembled by initializing a blank mask with original panoramic dimensions, then positioning each resized tooth mask using recorded cropping offsets and scaling factors.

Training was conducted in Python 3.10.13 using PyTorch 2.1.2 across different computational environments. The Hospital A dataset utilized Google Colaboratory Pro Plus with NVIDIA Tesla T4 GPU, while secondary dataset experiments employed NVIDIA Tesla V100 GPU infrastructure. Faster R-CNN employed Adam optimizer with combined loss function ($L_{Total} = L_{cls} + L_{box} + L_{obj} + L_{rpn-box}$), where $L_{cls}$ denotes classification loss for region of interest prediction, $L_{box}$ quantifies bounding box discrepancy, $L_{obj}$ measures object presence confidence, and $L_{rpn-box}$ corresponds to anchor box regression loss.

U-Net adopted AdamW optimizer with weight decay of 0.001 and Dice Loss:

$$Dice = \frac{2 \cdot TP}{2 \cdot TP + FP + FN} \quad (1)$$

where TP, FP, and FN represent true positives, false positives, and false negatives respectively.

Training hyperparameters included Faster R-CNN training for 100 epochs with batch size 4 and learning rate 0.001, while U-Net trained for 100 epochs with batch size 32 and learning rate 0.0001 (Table 2).

**Table 2.** Summary of the parameters applied to each model.

| Model | Faster R-CNN | U-Net |
|---|---|---|
| Batch size | 4 | 32 |
| Epoch | 100 | 100 |
| Optimizer | Adam | AdamW |
| Weight decay factor | - | 0.001 |
| Learning rate | 0.001 | 0.0001 |
| Loss function | Total loss | Dice loss |

Performance evaluation utilized four metrics: IoU, Dice coefficient, Recall, and Precision:

$$IoU = \frac{TP}{TP+FP+FN}\ ,\ Recall = \frac{TP}{TP+FN},\ Precision = \frac{TP}{TP+FP} \tag{2}$$

# 3 Results

## 3.1 Segmentation Results

The dual-stage framework achieved robust quantitative performance on the test dataset with IoU of 0.9013, Dice coefficient of 0.9482, Recall of 0.9433, and Precision of 0.9774. Training converged successfully with minimal overfitting, achieving stable performance after 20 epochs with final IoU values plateauing at approximately 0.965.

Fig. 5 demonstrates the U-Net model's segmentation performance on individual tooth images, comparing predicted outputs with ground truth annotations across both carious and healthy cases. For teeth containing carious lesions, the segmentation outputs closely aligned with expert-annotated masks, demonstrating high precision in identifying decay regions. The model successfully delineated lesion boundaries while preserving anatomical accuracy. For healthy tooth samples, the model correctly produced empty segmentation masks corresponding to the absence of pathological findings, indicating reliable discrimination between carious and non-carious dental tissue. Following individual tooth segmentation, all predicted masks were systematically integrated back into the panoramic radiograph format. This integration process enabled comprehensive visualization of lesion distribution across the entire dental arch, providing clinicians with a holistic perspective on the patient's oral health status.

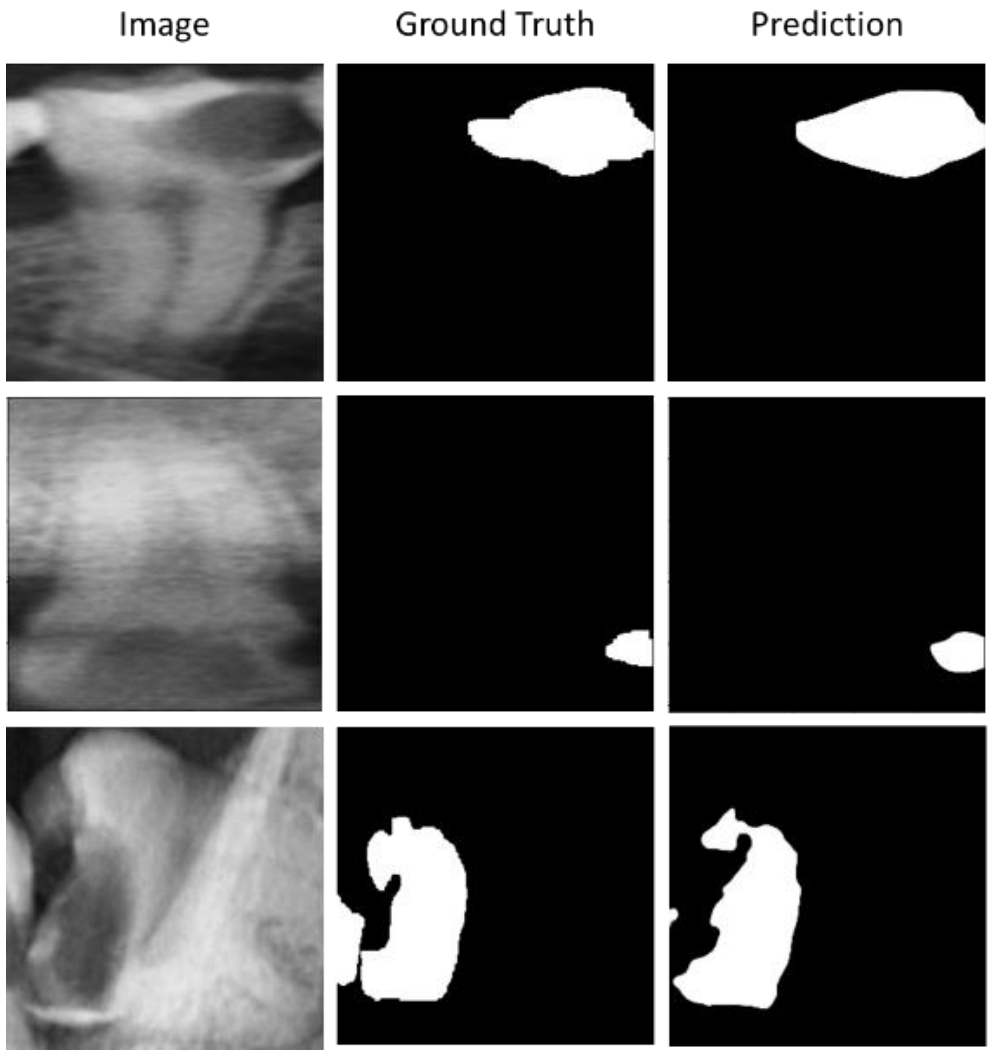

**Fig. 5**. Comparison of predictions and ground truth.

Fig. 6 provides magnified panoramic visualization with predicted carious regions overlaid in green for direct comparison with ground truth annotations. The first column displays unprocessed panoramic radiographs, while the second and third columns show manually annotated references and model predictions respectively. The magnified view clearly illustrates the model's capability to accurately pinpoint carious lesion locations across the dental arch while maintaining spatial coherence with the original anatomical structure. The visual results demonstrated the framework's effectiveness in both isolated tooth-level analysis and comprehensive panoramic assessment, supporting clinical applications requiring detailed lesion localization and extent evaluation.

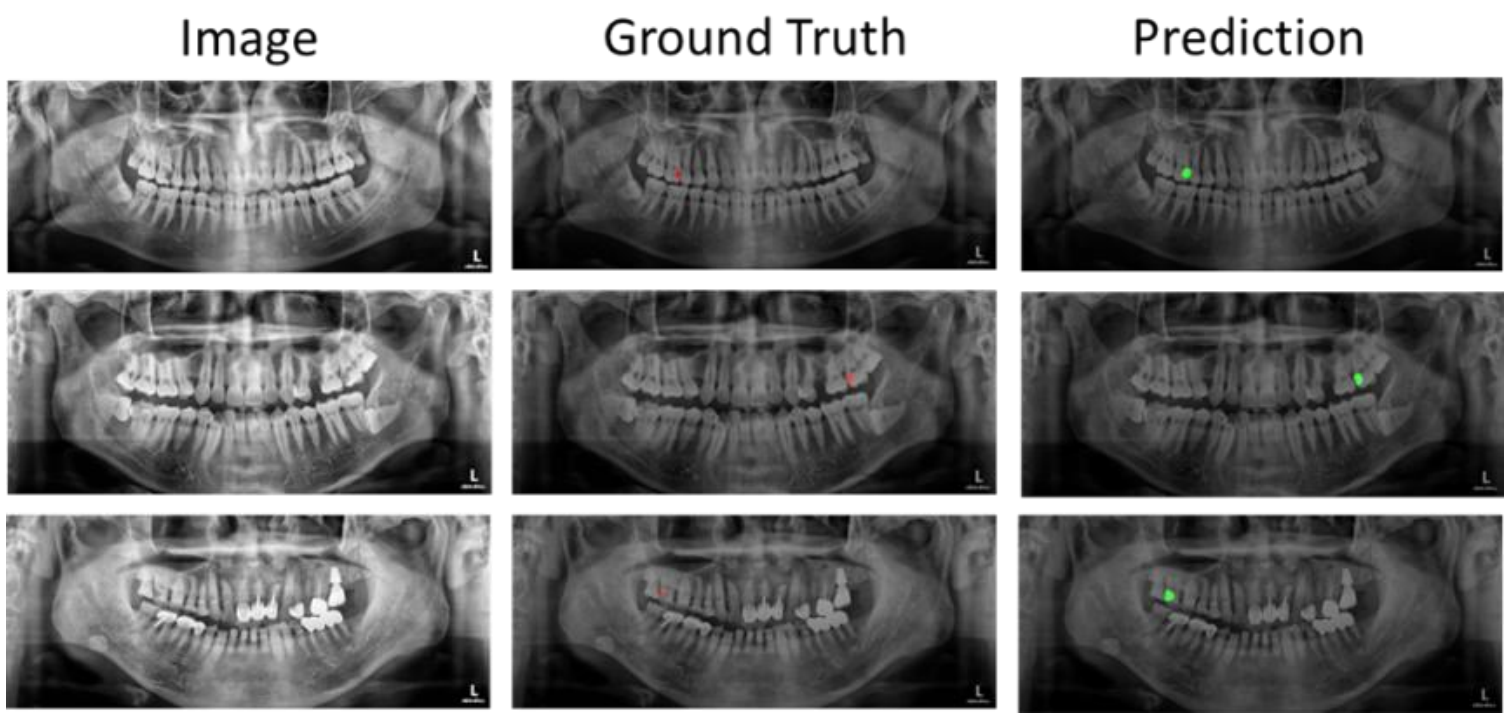

**Fig. 6**. Magnified panoramic image with a comparison of ground truth and prediction.

### 3.2 Comparative Analysis

Table 3 presents a comparative analysis of the proposed framework alongside three established methods: the Ensemble approach, the Custom U-Net, and the nnU-Net model. The Ensemble method showed moderate segmentation capability with limited detection of actual carious regions. The Custom U-Net demonstrated superior performance when trained on a larger dataset. The nnU-Net achieved perfect Precision but showed limitations in spatial overlap precision with a relatively lower Dice coefficient. The proposed method demonstrates superior overall performance across all metrics, effectively balancing precision and recall while achieving robust spatial alignment between predicted and ground truth annotations.

**Table 3.** Comparative performance analysis

| Methodology | Dataset | IoU | Dice | Recall | Precision |
|---|---|---|---|---|---|
| Ensemble [21] | 108 | 0.518 | 0.682 | 0.591 | - |
| Custom U-Net [22] | 1,159 | - | 0.936 | 0.860 | 0.941 |
| nnU-Net [23] | 1,160 | 0.785 | 0.663 | 0.821 | 1.00 |
| Proposed method | 18,131 | 0.901 | 0.948 | 0.943 | 0.977 |

## 4 Conclusions

This study presents a dual-stage deep learning framework for automated dental caries segmentation in panoramic radiographs, integrating Faster R-CNN for tooth localization with U-Net for precise lesion segmentation. Our integrated architecture successfully addresses the inherent challenges of panoramic imaging through a systematic two-stage approach that first identifies individual tooth regions and subsequently performs accurate lesion segmentation within these localized areas.

A critical innovation of this work lies in addressing the data annotation bottleneck that limits the scalability of deep learning models in dental diagnostics. Our systematic transformation pipeline converts large-scale polygon-annotated datasets into high-resolution binary segmentation masks, enabling pixel-wise supervised learning without the prohibitive costs of manual annotation. The dual-stage training strategy, which integrates both expert-verified and algorithmically-processed labels, successfully bridged the gap between annotation availability and model performance requirements. By converting 3,000 polygon-annotated images into 18,131 individual tooth-level training samples, our methodology substantially expands the utility of existing datasets for advanced segmentation model training.

Experimental evaluation demonstrates robust performance with IoU of 0.9013, Dice coefficient of 0.9482, Recall of 0.9433, and Precision of 0.9774, showing superior accuracy compared to existing methods while significantly reducing false positive rates. The high Precision achieved is particularly significant for clinical adoption, as it minimizes unnecessary interventions while maintaining strong detection sensitivity. These findings highlight the potential for implementing scalable, automated caries detection systems that can provide consistent, objective diagnostic support, reduce inter-clinician variability, and enhance treatment planning accuracy in diverse clinical environments. Future work will focus on expanding dataset diversity and validating clinical integration pathways to support widespread adoption of AI-assisted dental diagnostics in routine practice.